\documentclass{article}

\usepackage{PRIMEarxiv}

\usepackage[utf8]{inputenc} 
\usepackage[T1]{fontenc}    
\usepackage{hyperref}       
\usepackage{url}            
\usepackage{booktabs}       
\usepackage{amsfonts}       
\usepackage{nicefrac}       
\usepackage{microtype}      
\usepackage{lipsum}
\usepackage{fancyhdr}       
\usepackage{graphicx}       
\graphicspath{{media/}}     

\title{Automatic Essay Scoring in a Brazilian Scenario 

}

\author{
  Felipe Akio Matsuoka\\
  Faculdade de Ciências Médicas da Santa Casa de São Paulo \\
  São Paulo\\
  \texttt{felipe.matsuoka@aluno.fcmsantacasasp.edu.br} \\
}

\begin{document}
\maketitle

\begin{abstract}

This paper presents a novel Automatic Essay Scoring (AES) algorithm tailored for the Portuguese-language essays of Brazil's Exame Nacional do Ensino Médio (ENEM), addressing the challenges in traditional human grading systems. This approach leverages advanced deep learning techniques to align closely with human grading criteria, targeting efficiency and scalability in evaluating large volumes of student essays. This research not only responds to the logistical and financial constraints of manual grading in Brazilian educational assessments but also promises to enhance fairness and consistency in scoring, marking a significant step forward in the application of AES in large-scale academic settings.
\end{abstract}

\keywords{Automatic Essay Scoring \and Natural Language Processing \and Deep Learning \and Transformers}

\section{Introduction}
The evolution of educational assessment methods has been influenced by technological advancements, particularly in the realm of Automatic Essay Scoring (AES)\cite{ke2019automated}. This technology, leveraging the power of artificial intelligence, has emerged as a promising tool in evaluating written responses, especially in large-scale settings. The adoption of AES has become increasingly pertinent in countries like Brazil, where standardized tests play a crucial role in determining access to higher education. However, the journey towards integrating AES in such contexts has unique challenges and considerations.

One of the primary challenges faced in Brazil's educational assessment is the logistical and financial limitations associated with the traditional human grading system. With vast numbers of students participating in key examinations, the process of grading becomes not only time-consuming but also a significant financial burden on the educational system. This situation often leads to prolonged waiting periods.

Recognizing these challenges, this research introduces an automatic grading algorithm specifically designed for Portuguese-language essays. This algorithm harnesses the advancements in deep learning techniques. By doing so, not only addresses the linguistic nuances of the Portuguese language but also aligns closely with the criteria used by human graders.

In the context of Brazil, the Exame Nacional do Ensino Médio (ENEM) \footnote{\url{https://www.gov.br/inep/pt-br/areas-de-atuacao/avaliacao-e-exames-educacionais/enem}} stands as a main standardized test that significantly influences higher education for Brazilian students. Designed as a comprehensive assessment tool, ENEM evaluates a wide range of competencies spanning various academic disciplines. Among its various components, the essay section of the ENEM is particularly noteworthy. This segment of the test is designed to assess students' abilities to construct coherent arguments, demonstrate critical thinking, and exhibit proficiency in written communication in Portuguese. The essay topic typically revolves around contemporary social, political, or cultural issues, requiring students to engage with complex ideas and articulate their perspectives effectively.

The scoring of these essays has traditionally been the purview of human graders, who evaluate each response based on a set of predefined criteria. These criteria involves aspects like the development of ideas, coherence and cohesion in argumentation, command over language, and adherence to the proposed theme. Given the subjective nature of essay grading, achieving consistency and fairness in scores is a significant challenge. Moreover, with hundreds of thousands of students participating in the ENEM annually, the manual grading of essays is not only a labor-intensive and time-consuming process but also prone to inconsistencies.

This research, focuses on addressing these specific challenges posed by the ENEM essay component. By developing an AES algorithm tailored to the nuances of the Portuguese language and the specific grading criteria of the ENEM essay, this project aims to introduce a more efficient, consistent, and scalable approach to essay evaluation.

\section{Materials and Methods}
\label{sec:Materials_Methods}

For the all the data processing and model creation, it was opted to use Python language programming on its version 3.10 and the Google Colab platform. 

\subsection{Dataset Overview}

For the development of the algorithm, it was utilized the "Essay-br" dataset\cite{marinho-et-al-21}. This dataset comprises essays in Brazilian Portuguese, written by high school students for the ENEM (Exame Nacional do Ensino Médio). ENEM is a national standardized test in Brazil that determines eligibility for admission to various national universities. The dataset is publicly accessible on GitHub\footnote{\url{https://github.com/rafaelanchieta/essay}} and includes not only the essays but also Python scripts to help build the dataset and demonstrate how to read and process the essays.

The essays in the "Essay-br" dataset adhere to a structured dissertation-argumentative format. Each essay addresses a specific social issue, assigned to the students as the central theme. Students are expected to conduct an in-depth analysis of these social challenges and propose realistic, well-formulated solutions.

The grading criteria for these essays are segmented into five distinct categories, with each category being scored on a scale from 0 to 200, in 40-point increments. This structured approach to grading allows for a detailed assessment of the students' analytical and problem-solving skills, as well as their ability to effectively communicate their ideas.

\begin{table}[h]
\centering
\begin{tabular}{|c|p{10cm}|}
\hline
\textbf{Competence} & \textbf{Description} \\
\hline
1 & Adherence to the formal written norm of Portuguese. \\
\hline
2 & Conforming to the argumentative text genre and the proposed topic (theme), to develop a text, using knowledge from different areas. \\
\hline
3 & Selecting, relating, organizing, and interpreting data and arguments in defense of a point of view. \\
\hline
4 & Using argumentative linguistic structures. \\
\hline
5 & Elaborating a proposal to solve the problem in question. \\
\hline
\end{tabular}
\caption{Competences for Essay Evaluation}
\label{table:essay_competences}
\end{table}

\subsubsection{Exploratory Data Analysis}

The dataset comprises a total of 6,577 essays that have been evaluated, encompassing 151 distinct prompts. The cumulative scores, representing the sum of five individual grades, are presented visually in Figure 1. Each of the five grades is assigned within fixed intervals of 40 points. The distribution of these individual grades is systematically organized and can be reviewed in Table 2. 

The model was divided in training set (70\%), validation set(15\%) and holdout test set (15\%). 

\begin{table}[h]
\centering
\begin{tabular}{|c|c|c|c|c|c|}
\hline
Score & C1   & C2   & C3   & C4   & C5   \\ \hline
0     & 107  & 123  & 185  & 207  & 511  \\ 
40    & 24   & 93   & 164  & 65   & 297  \\ 
80    & 524  & 918  & 1606 & 885  & 1335 \\ 
120   & 3145 & 2446 & 3054 & 2459 & 2288 \\ 
160   & 2483 & 2425 & 1377 & 1822 & 1535 \\ 
200   & 294  & 572  & 191  & 1139 & 611  \\ \hline
\end{tabular}
\caption{Counts of grade values across competencies}
\label{tab:grade_counts}
\end{table}

\begin{figure}[h!]
\centering
\includegraphics[width=0.6\textwidth]{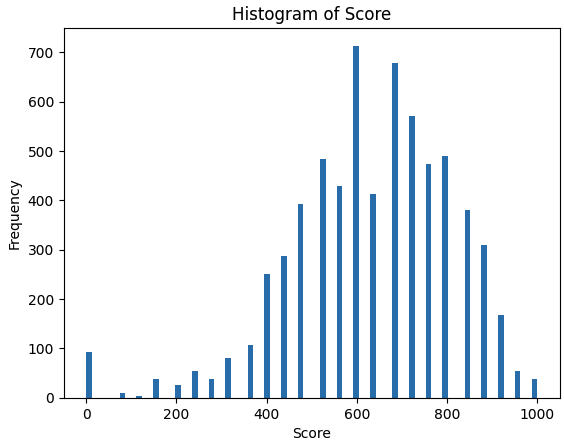}
\caption{Histogram of total grades}
\label{fig:histogram}
\end{figure}

\subsubsection{Preprocessing}
In the processing stage of the "Essay-BR" dataset, a significant improvement was implemented to better evaluate how the essay's content relates to its given theme or prompt.

We modified the input format for our model to include both the theme and the essay text. These two parts are separated by a special symbol '<SEP>', serving as a clear boundary between the theme and the essay response. This method allows the model to recognize and analyze the theme and essay as interconnected yet separate elements of the input. This enhancement aids in a more refined and precise evaluation of the essay's effectiveness in addressing the specified theme.

\subsection{Model Development}
To enhance essay grading efficiency, we developed a deep learning model based on the BERT (Bidirectional Encoder Representations from Transformers) architecture\cite{devlin2019bert}. Our specialized model, named BERT\_ENEM\_Regression, is tailored for regression tasks, such as assessing essays based on the five competencies set by ENEM.
\subsubsection{Model structure}
Key features of our model include:
\begin{enumerate}
    \item Pre-trained BERT Base: We utilize the BERTimbau \cite{inbook} a pre-trained BERT model adept at processing Portuguese texts. This base provides a robust foundation for understanding the nuances of the Portuguese language.
    \item Dropout Layer\cite{hinton2012improving}: To prevent overfitting — a common issue in machine learning models — we've included a dropout layer with a rate of 0.3. This helps the model generalize better to unseen data.
    \item Linear Layer for Competency Scoring: A crucial component is the linear layer, which aligns with BERT's hidden size (768 inputs) and has an output size of 5. This directly correlates to the five competencies in the ENEM grading system, allowing for precise score predictions for each essay.
\end{enumerate}

There are two versions of BERTimbau:
\begin{itemize}
    \item BERTimbau Base: Contains 12 layers with 110 million parameters
    \item BERTimbau Large: More complex with 24 layers and 335 million parameters
\end{itemize}

Both versions were trained on a large Portuguese corpus, BrWaC (Brazilian Web as Corpus)\cite{wagner-filho-etal-2018-brwac}, and are adept at handling the specificities of Brazilian Portuguese. Their state-of-the-art performance in key NLP tasks showcases their effectiveness in processing this language, making them invaluable for related NLP applications. In this work, we opted to use the BERTimbau base.

\subsubsection{Tokenization Process}
The BERTimbau model's tokenizer operates on the principles common to BERT tokenizers but is specifically trained on Brazilian Portuguese text. The tokenizer works by breaking down text into smaller units, known as tokens, and then converting these tokens into numerical IDs that the model can understand.

The tokenizer has a comprehensive vocabulary list that it references to convert words into tokens. When encountering words not in its vocabulary, the tokenizer applies a technique called WordPiece tokenization. This method allows the tokenizer to break down unknown or complex words into known subwords or smaller tokens. For instance, a word that is not in the vocabulary might be split into smaller segments with recognizable prefixes or suffixes, which are then processed individually.

Each input string is processed by adding special tokens that mark the beginning and end of the string or sentence. For BERT, these are the `[CLS]` and `[SEP]` tokens, which are crucial for the model to understand the context and separation of sentences.

Additionally, the tokenizer is designed to handle punctuation by adding whitespace around punctuation characters. This allows the tokenizer to treat them as separate tokens, ensuring that the model can distinguish between words and punctuation during processing.

When dealing with multiple sentences or pairing questions with context, the tokenizer uses `token\_type\_ids` to differentiate between the two sequences. Moreover, `attention\_mask` is used to manage varying sentence lengths, especially when processing batches of text data. It ensures that the model pays attention to the actual content and ignores any padding introduced to equalize sentence lengths.

The tokenizer also includes functionality for handling out-of-vocabulary words using special tokens like `[UNK]` for unknown words, and it employs subword tokenization for a more effective and granular understanding of language.

In summary, the BERTimbau tokenizer is a powerful tool for preparing Brazilian Portuguese text for processing by the BERTimbau model, effectively managing the nuances of the language, and ensuring that the data is appropriately formatted for the best performance of the model in various NLP tasks. The details of the tokenizer's operation reflect the sophisticated capabilities of the BERT framework and its adaptations for specific languages like Brazilian Portuguese.

\section{Results}

The training of the model was conducted over 5 epochs, utilizing a batch size of 16. This process was carried out on an A6000 GPU, and the AdamW\cite{loshchilov2019decoupled} optimizer was employed for optimization.

The tables provided display performance metrics for different models applied to essay scoring, as is detailed in the Essay-BR dataset paper. Table 3 shows the Quadratic Weighted Kappa (QWK) scores for each model across five criteria (C1 to C5) and a total score. The QWK measures (Fig. 2) the agreement between the scores assigned by the automated system and those given by human raters, accounting for the possibility of chance agreement. Higher QWK scores indicate a closer match to human scoring. According to this table, the BERT\_ENEM\_Regression model shows superior performance with a total QWK of 0.79, indicating a strong correlation with human judgment across all criteria.

Table 4 presents the Root Mean Squared Error (RMSE) (Fig.3 ) for the same models and criteria. RMSE is a standard way to measure the error of a model in predicting quantitative data. Lower RMSE values represent more accurate predictions. Here again, the BERT\_ENEM\_Regression model outperforms the others with a total RMSE of 90.96, suggesting it has the lowest prediction error and thus is the most accurate among the listed models.

These metrics give a quantitative view of how the proposed algorithm performs, demonstrating its effectiveness in automatic essay scoring within the Brazilian educational context, as observed with the Essay-BR dataset.

\begin{figure}[h]
\label{qwk}
\centering
\[
\kappa = 1 - \frac{\sum_{i,j}w_{ij}o_{ij}}{\sum_{i,j}w_{ij}e_{ij}}
\]
\caption{The Quadratic Weighted Kappa (QWK) measures the agreement between two raters. Here, $w_{ij}$ is the weight for the disagreement between the $i$-th and $j$-th category, $o_{ij}$ is the observed agreement, and $e_{ij}$ is the expected agreement under chance.}
\end{figure}

\begin{figure}[h]
\label{rmse}
\centering
\[
RMSE = \sqrt{\frac{1}{n}\sum_{i=1}^{n}(y_i - \hat{y_i})^2}
\]
\caption{The Root Mean Squared Error (RMSE) quantifies the difference between predicted values $\hat{y_i}$ and observed values $y_i$, averaged over $n$ observations.}
\end{figure}
In orded to evaluate the models performance, after runniing the model's inference on the holdout test set,  achieved the results are shown on tables \ref{qwk_table} and 4. Besides that, to give more detail on the model's perfomance the results were compared against the same research works shown in the "Essay-br" dataset

\begin{table}[h!]
\label{qwk_table}
\centering
\caption{Quadratic Weighted Kappa (QWK)}
\begin{tabular}{lccccc|c}
\hline
\textbf{Model} & \textbf{C1} & \textbf{C2} & \textbf{C3} & \textbf{C4} & \textbf{C5} & \textbf{Total} \\
\hline
Amorim and Veloso 2017\cite{amorim-veloso-2017-multi} & 0.39 & 0.46 & 0.40 & 0.38 & 0.34 & 0.49 \\
Fonseca et al. 2018\cite{fonseca} & 0.44 & 0.48 & 0.42 & 0.47 & 0.38 & 0.53 \\
BERT\_ENEM\_Regression & 0.74 & 0.78 & 0.76 & 0.84 & 0.79 & 0.79 \\
\hline
\end{tabular}
\end{table}

\begin{table}[h!]
\label{rmse_table}
\centering
\caption{Rooted Mean Squared Error (RMSE)}
\begin{tabular}{lccccc|c}
\hline
\textbf{Model} & \textbf{C1} & \textbf{C2} & \textbf{C3} & \textbf{C4} & \textbf{C5} & \textbf{Total} \\
\hline
Amorim and Veloso 2017 \cite{amorim-veloso-2017-multi} & 32.26 & 33.45 & 38.2 & 39.35 & 48.18 & 161.09 \\
Fonseca et al. 2018 \cite{fonseca}& 32.16 & 33.55 & 38 & 38.32 & 47.66 & 157.33 \\
BERT\_ENEM\_Regression & 21.77 & 24.37 & 25.30 & 24.15 & 34.03 & 90.96 \\
\hline
\end{tabular}
\end{table}

\section{Discussion}

The BERT\_Regression\_Base model shows exceptional performance, likely due to its advanced design and deep language understanding, giving it an advantage in assessing essay quality. Its success underscores the capability of sophisticated tools like BERT in grading a large number of essays, vital in educational environments.

However, the model scores lower in Competence 1, which evaluates grammatical accuracy. This may be because BERT's tokenization, which breaks text into smaller pieces, might overlook some grammatical errors. These errors get 'smoothed out' in the process, making it hard for the model to spot and mark them correctly. As a result, it scores higher in Root Mean Square Error (RMSE) and lower in Quadratic Weighted Kappa (QWK) for Competence 1. This suggests that while BERT is effective for general text understanding, it may not be as accurate in judging grammatical quality, pointing to an area for improvement, particularly where grammatical precision is key.

Additionally, the dataset's grade distribution, mostly skewed towards higher scores, limits the analysis and model training. This skewness could cause the model to be better at identifying and predicting higher grades but less accurate with lower-scoring essays. This limitation occurs because the model gets less exposure to a range of lower-grade examples during training, affecting its ability to generalize across all score ranges. This issue is important as it impacts the model's real-world effectiveness in educational settings, where a balanced assessment across various competencies is essential. Future improvements could include diversifying the dataset or adjusting the model to better recognize under-represented score ranges.

In summary, while deep learning \cite{tashu2022deep}, particularly transformer models like BERT shows promise in Automated Essay Scoring (AES), challenges remain. These include the model's difficulty with grammatical accuracy and potential biases due to uneven data distribution. Contrastingly, traditional methods using handcrafted features, based on linguistic and stylistic analysis, can be effective but might not fully capture the nuanced understanding of language and context like a human or deep learning model.

\bibliographystyle{unsrt}  
\bibliography{references}

\begin{thebibliography}{10}

\bibitem{ke2019automated}
Zixuan Ke and Vincent Ng.
\newblock Automated essay scoring: A survey of the state of the art.
\newblock In {\em IJCAI}, volume~19, pages 6300--6308, 2019.

\bibitem{marinho-et-al-21}
Jeziel Marinho, Rafael Anchiêta, and Raimundo Moura.
\newblock Essay-br: a brazilian corpus of essays.
\newblock In {\em Anais do III Dataset Showcase Workshop}, pages 53--64, Online, 2021. Sociedade Brasileira de Computação.

\bibitem{devlin2019bert}
Jacob Devlin, Ming-Wei Chang, Kenton Lee, and Kristina Toutanova.
\newblock Bert: Pre-training of deep bidirectional transformers for language understanding, 2019.

\bibitem{inbook}
Fábio Souza, Rodrigo Nogueira, and Roberto Lotufo.
\newblock {\em BERTimbau: Pretrained BERT Models for Brazilian Portuguese}, pages 403--417.
\newblock 10 2020.

\bibitem{hinton2012improving}
Geoffrey~E. Hinton, Nitish Srivastava, Alex Krizhevsky, Ilya Sutskever, and Ruslan~R. Salakhutdinov.
\newblock Improving neural networks by preventing co-adaptation of feature detectors, 2012.

\bibitem{wagner-filho-etal-2018-brwac}
Jorge~A. Wagner~Filho, Rodrigo Wilkens, Marco Idiart, and Aline Villavicencio.
\newblock The br{W}a{C} corpus: A new open resource for {B}razilian {P}ortuguese.
\newblock In Nicoletta Calzolari, Khalid Choukri, Christopher Cieri, Thierry Declerck, Sara Goggi, Koiti Hasida, Hitoshi Isahara, Bente Maegaard, Joseph Mariani, H{\'e}l{\`e}ne Mazo, Asuncion Moreno, Jan Odijk, Stelios Piperidis, and Takenobu Tokunaga, editors, {\em Proceedings of the Eleventh International Conference on Language Resources and Evaluation ({LREC} 2018)}, Miyazaki, Japan, May 2018. European Language Resources Association (ELRA).

\bibitem{loshchilov2019decoupled}
Ilya Loshchilov and Frank Hutter.
\newblock Decoupled weight decay regularization, 2019.

\bibitem{amorim-veloso-2017-multi}
Evelin Amorim and Adriano Veloso.
\newblock A multi-aspect analysis of automatic essay scoring for {B}razilian {P}ortuguese.
\newblock In Florian Kunneman, Uxoa I{\~n}urrieta, John~J. Camilleri, and Mariona~Coll Ardanuy, editors, {\em Proceedings of the Student Research Workshop at the 15th Conference of the {E}uropean Chapter of the Association for Computational Linguistics}, pages 94--102, Valencia, Spain, April 2017. Association for Computational Linguistics.

\bibitem{fonseca}
Erick Fonseca, Ivo Medeiros, Dayse Kamikawachi, and Alessandro Bokan.
\newblock {\em Automatically Grading Brazilian Student Essays: 13th International Conference, PROPOR 2018, Canela, Brazil, September 24–26, 2018, Proceedings}, pages 170--179.
\newblock 01 2018.

\bibitem{tashu2022deep}
Tsegaye~Misikir Tashu, Chandresh~Kumar Maurya, and Tomas Horvath.
\newblock Deep learning architecture for automatic essay scoring, 2022.

\end{thebibliography}

\end{document}